\documentclass[conference]{IEEEtran}
\IEEEoverridecommandlockouts
\usepackage{cite}
\usepackage{amsmath,amssymb,amsfonts}
\usepackage{algorithmic}
\usepackage{graphicx}
\usepackage{textcomp}
\usepackage{hyperref}
\usepackage{cleveref}
\usepackage{xcolor}
\usepackage{footmisc}

\usepackage{listings}

\usepackage{multirow}
\usepackage{booktabs}
\usepackage{makecell}
\usepackage{amsfonts}
\usepackage{amsmath}
\usepackage{tablefootnote}
\usepackage{subcaption}
\usepackage{arydshln}

\def\BibTeX{{\rm B\kern-.05em{\sc i\kern-.025em b}\kern-.08em
    T\kern-.1667em\lower.7ex\hbox{E}\kern-.125emX}}
\begin{document}

\title{Keyword-Oriented Multimodal Modeling for Euphemism Identification}

\author {
    Yuxue Hu\textsuperscript{\rm 1,\rm 2,\rm 3,\rm 4*\thanks{* Equal contribution. \dag  ~Corresponding author.}},
    Junsong Li\textsuperscript{\rm 1*},
    Meixuan Chen\textsuperscript{\rm 1*},
    Dongyu Su\textsuperscript{\rm 1, \rm 2},
    Tongguan Wang\textsuperscript{\rm 1, \rm 2},
    Ying Sha\textsuperscript{\rm 1,\rm 2,\rm 3,\rm 4 \dag}\\
    \textit{\textsuperscript{\rm 1} College of Informatics, Huazhong Agricultural University, Wuhan, China}\\
    \textit{\textsuperscript{\rm 2}Key Laboratory of Smart Farming for Agricultural Animals, Wuhan, China}\\
    \textit{\textsuperscript{\rm 3}Hubei Engineering Technology Research Center of Agricultural Big Data, Wuhan, China}\\
    \textit{\textsuperscript{\rm 4} Engineering Research Center of Intelligent Technology for Agriculture, Ministry of Education}\\
    {\small \texttt{\{hyx, shaying\}@mail.hzau.edu.cn}, \texttt{\{wang\_tg,su\_dy\}@webmail.hzau.edu.cn}}\\
    {\small \texttt{\{isjunsong.li, meixuanchen419\}@gmail.com}}
}

\maketitle

\begin{abstract}
Euphemism identification deciphers the true meaning of euphemisms, such as linking “weed” (euphemism) to “marijuana” (target keyword) in illicit texts, aiding content moderation and combating underground markets. While existing methods are primarily text-based, the rise of social media highlights the need for multimodal analysis, incorporating text, images, and audio. However, the lack of multimodal datasets for euphemisms limits further research. To address this, we regard euphemisms and their corresponding target keywords as keywords and first introduce a keyword-oriented multimodal corpus of euphemisms (KOM-Euph), involving three datasets (Drug, Weapon, and Sexuality), including text, images, and speech. We further propose a keyword-oriented multimodal euphemism identification method (KOM-EI), which uses cross-modal feature alignment and dynamic fusion modules to explicitly utilize the visual and audio features of the keywords for efficient euphemism identification. Extensive experiments demonstrate that KOM-EI outperforms state-of-the-art models and large language models, and show the importance of our multimodal datasets \footnotemark\footnotetext{https://github.com/DHZ68/KOM-EI}.

\end{abstract}

\begin{IEEEkeywords}
Euphemism identification, multimodal, alignment, fusion
\end{IEEEkeywords}

\section{Introduction}
Euphemisms are indirect words or phrases used to replace harsh expressions, playing a significant role in linguistic communication. They are widely used on social media and darknet marketplaces to evade supervision \cite{yuan2018reading, takuro2020codewords, foye2021illicit}. For instance, “ice” and “weed” in Table \ref{table:eg1} are substitutes for “methamphetamine” and “marijuana”. These euphemisms can be vague, making it challenging to trace illegal transactions. Identifying the target keyword of a euphemism, i.e., euphemism identification, is crucial for improving content moderation and combating underground trading. However, euphemisms evolve like a “treadmill” \cite{pinker2003blank}, complicating the maintenance of an up-to-date corpus. Additionally, euphemisms can be used literally or figuratively, adding further complexity to the task.

Current methods detect euphemistic word usage, evolving from traditional NLP \cite{yuan2018reading, magu2018determining, lee2022searching} to deep learning models \cite{zhu2021self, zhu2021euphemistic, seethappan2022comparative}. However, these methods can only detect euphemisms, not identify their corresponding target keywords. Existing studies use self-supervised learning to label datasets but rely solely on textual context, ignoring multimodal semantic information crucial for euphemism evolution.

\begin{table}[t!]
    \caption{Examples of sentences containing euphemisms.} 
    \label{table:eg1}
    \centering
    \begin{tabular}{p{7.9cm}}
        \toprule
        \textbf{Example sentences }(euphemisms are in bold)\\
        \midrule
        1. We had already paid \$70 for some shitty \textbf{weed} from a taxi driver but we were interested in some \textbf{coke} and the cubans.\\
        2. For all vendors of \textbf{ice}, it seems pretty obvious that it is not as pure as they market it.\\	
        3. Back up before I pull my \textbf{nine} on you.\\
        \bottomrule
    \end{tabular}
    \vspace{-4mm} 
\end{table}

\begin{figure}[t!]
    \centering 
    \includegraphics[width=0.945\linewidth]{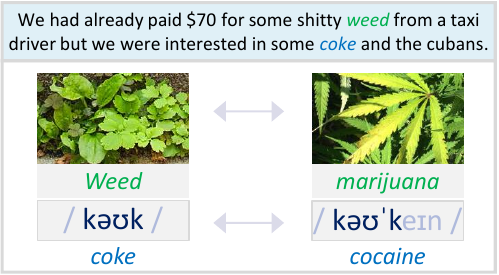} 
    \caption{Image and speech examples of keywords.} 
    \label{fig:1} 
    \vspace{-6mm} 
\end{figure}

Euphemisms in language evolution often arise from homophones, abbreviations, and image mappings \cite{ji2018creative}. As shown in Fig.~\ref{fig:1}, “weed” (a plant) and its euphemistic meaning (marijuana) share a semantic link to plants, supported by visual cues. Similarly, “Coke” euphemistically refers to cocaine due to its historical inclusion in the beverage and phonetic similarity. Language is recorded through text, but visual and audio modalities offer additional insights. 
Text is just one modality for recording language; visual and audio modalities provide additional information, illustrating language evolution. Additionally, some harmful content, such as drug abuse, utilizes the cross-modal characteristics of euphemisms on certain platforms to evade supervision \cite{rutherford2022getting,fuller2024understanding}. Moreover, using other modalities to complement text is effective in NLP tasks \cite{wang2022cross, yang2023confede, zeng2023explainable}. Thus, integrating multimodal data is crucial for euphemism identification. However, research remains text-focused, with the lack of multimodal data limiting progress.

To address these limitations, we create the first \textbf{K}eyword-\textbf{O}riented \textbf{M}ultimodal \textbf{Euph}emism dataset (KOM-Euph) based on the text-only datasets \cite{zhu2021self}, including text-image-speech triplets without labels. Recognizing that euphemisms often hinge on specific keywords carrying nuanced meanings across different modalities, we introduce a keyword-oriented approach. This method captures subtle semantic nuances, expanding understanding from mono- to multi-modality and improving automatic identification through multimodal analysis.

Additionally, to better utilize the multimodal information of euphemisms from text, vision, and audio, we propose a \textbf{K}eyword-\textbf{O}riented \textbf{M}ultimodal \textbf{E}uphemism \textbf{I}dentification method (KOM-EI). KOM-EI generates comprehensive semantics of euphemisms by explicitly using visual and audio features. It employs feature alignment to align cross-modal features through contrastive learning and uses dynamic feature fusion to flexibly obtain cross-modal features via cross-attention and gated units. This approach enhances the model's ability to exploit text, vision, and audio features, leading to more accurate identification. Experiments show that our method achieves top-1 identification accuracies 45-60\% higher than state-of-the-art baseline methods. 


Our contributions are as follows:
\begin{itemize}
\item To the best of our knowledge, we are the first that contribute a novel keyword-oriented multimodal euphemism corpus (KOM-Euph) with 86K text-image-speech triplets involving three domains.
\item We propose a keyword-oriented multimodal fusion method, using cross-modal feature alignment and dynamic fusion to explicitly exploit the text-image-speech features to identify euphemisms.

\item Extensive experiments on KOM-Euph show that our model builds new state-of-the-art performance that beats large language models and demonstrates the importance of our datasets.
\end{itemize}


\begin{figure*}[t!]
	\centering
	\includegraphics[width=1\textwidth]{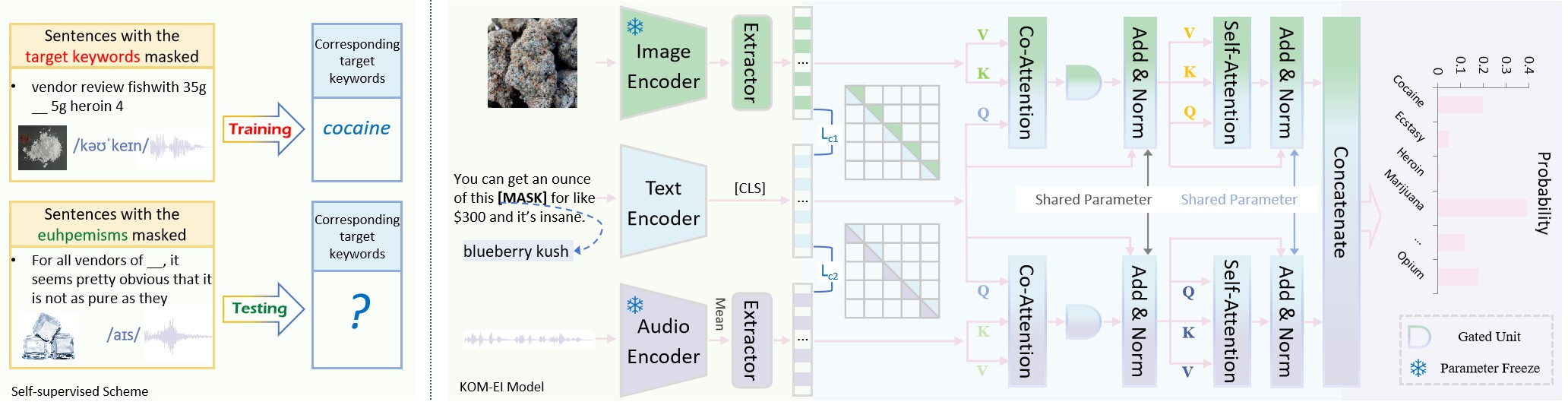}
\caption{The left part illustrates the self-supervised learning scheme for constructing labeled training sets, where sentences with masked target keywords are labeled and enriched with multimodal information. The right part shows the architecture of our KOM-EI, which consists of three modules: (1) Feature Representation extracts text, image, and speech features using pre-trained models; (2) Feature Fusion dynamically aligns and integrates multimodal features via co-attention mechanisms; and (3) Prediction identifies target keywords based on fused features.}
\label{fig:2}
\vspace{-3mm}
\end{figure*}

\section{Methodology}
\subsection{Task Definition}

This study focuses on identifying the target keyword \( t_j \) corresponding to a euphemism \( euph \) in a sentence \( s = [w_1, \dots, w_i, euph, \dots, w_m] \), where \( s \in \text{Set} \), and \(\text{Set}\) is the collection of masked sentences. Given sentences \( S \), target keywords $T=\{t_1,...,t_j,...,t_n\}$ , and associated images and speech, the goal is to determine, e.g., that ``ice'' means ``methamphetamine'' and ``nine'' means ``gun'' (Table \ref{table:eg1}).


\subsection{Framework}
We propose KOM-EI, a keyword-oriented multimodal euphemism identification method integrating text, visual, and audio modalities for improved accuracy. Following \cite{zhu2021self}, we adopt a self-supervised scheme where sentences with masked target keywords serve as inputs and their actual target keywords as labels. During testing, the model predicts target keywords for sentences with masked euphemisms. Unlike text-only methods, KOM-EI incorporates multimodal features during both training and testing, enhancing performance (Fig.~\ref{fig:2}).


\subsection{Feature Representation Module}
Euphemisms are often identified through contextual analysis, but relying solely on context can lead to ambiguity. Similar contexts for different euphemisms may confuse models, resulting in misidentification. For instance, in the sentence, “We had already paid \$70 for some shitty weed ... we were interested in some coke and the cubans,” distinguishing “weed” from “coke” is challenging with only sentence-level context.

Research suggests that the visual and audio aspects of a euphemism's literal meaning often relate to its implicit meaning. For example, both the literal and euphemistic meanings of “weed” (marijuana) involve plants visually, while “coke” is linked to “cocaine” due to the original drink’s ingredients and phonetics. Inspired by this, we incorporate multimodal information—text, vision, and audio—to extract semantically rich features for more accurate euphemism identification.



\textbf{Text Encoder}. Given BERT's effectiveness in capturing contextual semantics~\cite{devlin2019bert}, we employ a BERT model pre-trained on a euphemism corpus. For a masked sentence $ s = [\texttt{[CLS]}, w_1, \ldots, \texttt{[MASK]}, \ldots, w_m, \texttt{[SEP]}] $, where \texttt{[CLS]} and \texttt{[SEP]} denote boundary markers and \texttt{[MASK]} replaces the euphemism. We derive the textual representation as:
\begin{equation}\label{eq:1}
T = \mathrm{CLS\_BERT}(s),
\end{equation}
where $T \in \mathbb{R}^{d_g}$ is the global textual embedding.

\textbf{Image Encoder}. To obtain semantically rich image embeddings, we use a pre-trained CLIP model~\cite{radford2021learning}, known for bridging vision and language. To retain the model’s pretrained knowledge, we freeze its parameters and introduce a nonlinear projection layer:
\begin{equation}\label{eq:2}
    \hat{I} = \mathrm{CLIP}(\mathrm{Image}),
\end{equation}
\begin{equation}\label{eq:3}
    I = \mathrm{ReLU}(W_I \hat{I} + b_I),
\end{equation}
\noindent where $\hat{I} \in \mathbb{R}^{d_v}$ is the initial image embedding and $I \in \mathbb{R}^{d_g}$ is the projected embedding aligned to the common feature space.

\textbf{Speech Encoder}. To capture fine-grained acoustic characteristics, we employ Wav2Vec 2.0~\cite{NEURIPS2020_92d1e1eb} as the speech encoder. By freezing its pretrained weights and integrating an additional extractor, we obtain a global speech representation:
\begin{equation}\label{eq:4}
    \tilde{S} = \mathrm{Wav2Vec2}(\mathrm{Speech}) = [z_1, z_2, \ldots, z_T],
\end{equation}
\begin{equation}\label{eq:5}
    \hat{S} = \mathrm{Mean}(\tilde{S}),
\end{equation}
\begin{equation}\label{eq:6}
    S = \mathrm{ReLU}(W_S \hat{S} + b_S),
\end{equation}
\noindent where each $z_j \in \mathbb{R}^{d_s}$ represents the $j$-th time-step embedding, $\mathrm{Mean}(\cdot)$ denotes the average operation over time, and $S \in \mathbb{R}^{d_g}$ is the resulting global speech embedding.

\subsection{Dynamic Feature Fusion Module}
By integrating multimodal signals, euphemism identification benefits from complementary visual and auditory cues in addition to text. Yet, these modalities may also introduce extraneous information. Thus, we anchor on textual features and dynamically incorporate relevant elements from images and audio. We first use cross-modal contrastive learning to align text-image and text-speech embeddings, then apply cross-attention to extract complementary signals. Finally, a gated unit filters redundant inputs, refining the fused representation.


\textbf{Cross-modal Feature Alignment (CFA)}. Prior studies highlight modality gaps in multimodal models \cite{xu2021videoclip,liang2022mind,zhang2022contrastive}. CFA mitigates these gaps by aligning heterogeneous features. Given a sentence with a keyword, we pair it with the corresponding image or audio. Positive samples match keyword-text pairs with relevant modalities, while negative samples use mismatched pairs. The cross-modal contrastive loss promotes semantic alignment and discourages irrelevant associations.
\begin{equation}\label{eq:7}
\begin{aligned}
    L_{TI} = {-}&\sum_{i=1}^{|Set|}\sum_{j=1}^{|B|}\mathbb{I}([mask]_{i}=keyword_{j})\\
    &\log\frac{e^{sim(T_{i},I_{j})/\tau}}{\sum_{k=1}^{|B|}e^{sim(T_{i},I_{k})/\tau}},
\end{aligned}
\end{equation}
\begin{equation}\label{eq:8}
\begin{aligned}
    L_{TS} = {-}&\sum_{i=1}^{|Set|}\sum_{j=1}^{|B|}\mathbb{I}([mask]_{i}=keyword_{j})\\
    &\log\frac{e^{sim(T_{i},S_{j})/\tau}}{\sum_{k=1}^{|B|}e^{sim(T_{i},S_{k})/\tau}},  
\end{aligned}
\end{equation}

\noindent where $|B|$ is the batch size, $\mathbb{I}$ is an indicator, $[mask]_{i}$ is the keyword in $s$, $keyword_{j}$ refers to the keyword from the image or speech, $sim(\cdot,\cdot)$ is the cosine similarity, and $\tau$ is a temperature hyper-parameter.

\textbf{Cross-modal Attention (CA)}. To extract complementary information from other modalities, we use contextual features as anchors and apply cross-attention to highlight relevant data. Specifically, the query \( Q \) is linearly projected from textual features \( T \), while the key \( K \) and value \( V \) come from visual \( I \) or audio features \( S \):
\( Q = T W_q, K = I W_k / S W_k, V = I W_v / S W_v, Q/K/V \in \mathbb{R}^{d_g} \). CA is then applied to derive the context-queried visual features \( M_{TI} \) and audio features \( M_{TS} \).
\begin{equation}\label{eq:9}
\begin{aligned}
    M_{TI} &= \textrm{CA}(Q_{TI}, K_{TI}, V_{TI}),\\
    M_{TS} &= \textrm{CA}(Q_{TS}, K_{TS}, V_{TS}).
\end{aligned}
\end{equation}
\begin{figure*}[!ht] 
    \centering %
    \includegraphics[width=1\linewidth]{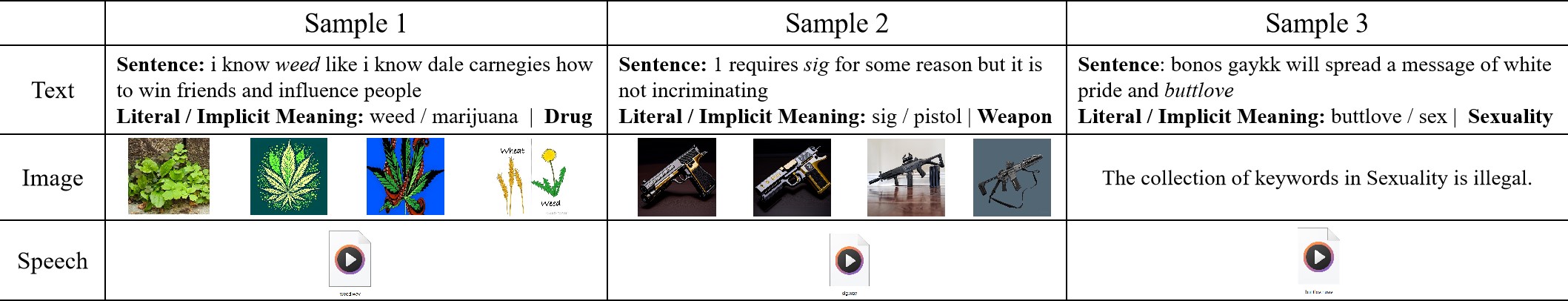} 
    \caption{Samples of multimodal datasets.} %
    \label{figure1} %
    \vspace{-2mm}    
\end{figure*}

\textbf{Gated Unit (GU)}. The GU filters the noise from visual or audio features, learning dynamic text-image and text-speech co-attention. This yields text-guided outputs $\hat{M_{TI}}$ and $\hat{M_{TS}}$, followed by an \textbf{Addition and Normalization} layer $\textrm{AN}_{\text{GU}}$:
\begin{equation}\label{eq:10}
\begin{aligned}
    \textrm{R}(X)&= \textrm{ReLU}(W_R X + b_R),\\
    \textrm{GU}(X)&=\sigma(W_G\textrm{R}(X) + b_G)\cdot X,
\end{aligned}
\end{equation}
\begin{equation}\label{eq:11}
    \tilde{M_{TI}} = \textrm{GU}(M_{TI}),\;\tilde{M_{TS}} = \textrm{GU}(M_{TS}),
\end{equation}
\begin{equation}\label{eq:12}
\begin{aligned}
    \hat{M_{TI}} &= \textrm{AN}_{\text{GU}}(\tilde{M_{TI}} + {M_{TI}}),\\
    \hat{M_{TS}} &= \textrm{AN}_{\text{GU}}(\tilde{M_{TS}} + {M_{TS}}).    
\end{aligned}
\end{equation}

Next, we employ a \textbf{Self-Attention (SA)} layer followed by an AN layer $\textrm{AN}_{\text{SA}}$ to refine the text-guided output $\hat{M_{TI}}$. $\hat{Q_{TI}} = \hat{M_{TI}}W_{q^{TI}},
\hat{K_{TI}} = \hat{M_{TI}}W_{k^{TI}}, \hat{V_{TI}} = \hat{M_{TI}}W_{v^{TI}}$.
\begin{equation}\label{eq:13}
    \widehat{M_{TI}} = \textrm{SA}(\hat{Q_{TI}}, \hat{K_{TI}}, \hat{V_{TI}}),
\end{equation}
\begin{equation}\label{eq:14}
    \overline{M_{TI}} = \textrm{AN}_{\text{SA}}(\widehat{M_{TI}} + \hat{M_{TI}}).
\end{equation}

Similarly, we can get the enhanced features $\overline{M_{TS}}$. Finally, the dynamic fusion features are obtained as follows:
\begin{equation}\label{eq:15}
        H(s) = W_H(\overline{M_{TI}};\overline{M_{TS}})) + b_H,
\end{equation}
\noindent where $W_H \in R^{d_g \times 2d_g}$, $b_H \in R^{d_g}$ are the model parameters, and (;) means concatenation.

\subsection{Prediction Module}
After obtaining the dynamic fusion feature \( H(s) \), a classifier identifies the target keyword for a masked sentence. The probability is given by:
\begin{equation}\label{eq:16}
	P(t_j|s) = \text{softmax}(W(h(t_j)\odot H(s)) + b),
\end{equation}
where \(W \in \mathbb{R}^{d_g}, b \in \mathbb{R}\) are parameters, \(\odot\) denotes element-wise multiplication, and \(h(t_j)\) is the learned representation of the target keyword’s class label. The loss is defined as:
\begin{equation}\label{eq:17}
	L_P = -\sum_{j=1}^{n}H_g \log P(t_j|s),
\end{equation}
where \(n\) is the number of categories and \(H_g\) is the one-hot ground-truth vector. Within a category, target keywords of the same subcategory share identical meanings.

\subsection{Training and Inference}
With the prediction loss \(L_P\) and the alignment losses \(L_{TI}\) and \(L_{TS}\), the final objective is:
\begin{equation}\label{eq:101}
	J = \alpha L_P + \beta L_{TI} + \gamma L_{TS},
\end{equation}
where \(\alpha,\beta,\gamma\) balance these terms. During inference, only the main prediction task is used, without auxiliary alignment.

\section{Experiments and Analysis}
We evaluate KOM-EI on the KOM-Euph corpus and compare its performance against baseline models.

\subsection{KOM-Euph Dataset}
To address the lack of multimodal resources for euphemisms, we construct KOM-Euph, the first dataset inspired by the role of visual and audio cues in euphemism evolution.


\paragraph{Data Construction} Building on the text-only Euph corpus \cite{zhu2021self}, derived from Reddit, Gab, and Slangpedia, we introduce multimodal elements. Euph includes Drug, Weapon, and Sexuality domains, with audio added for Sexuality due to legal constraints and both audio and visual modalities for Drug and Weapon.


\textbf{Visual Modality Construction}. We collect keyword images from Google, Wikipedia, and the Kandinsky 2.2 model\footnote{\url{https://github.com/ai-forever/Kandinsky-2}}, retrieving 10 images each from Google and Wikipedia, and generating 5 images via Kandinsky, for a total of 25 images per keyword. We hire a linguistics expert to guide 6 undergraduates in selecting the top 4 representative images. For ambiguous keywords (e.g., “k4”, “404”), we select 2 top-ranked images from Google and 2 generated images, ensuring that each keyword has a curated set of high-quality visuals.

\textbf{Audio Modality Construction}. We generate a single standard pronunciation clip per keyword using Bark\footnote{\url{https://github.com/suno-ai/bark}\label{foot:a5}}, without prosodic variations. This provides a consistent, high-quality acoustic reference to aid euphemism identification.

\paragraph{Dataset Statistics}
Table~\ref{table:2} summarizes the datasets: Drug, Weapon, and Sexuality contain 33, 9, and 12 subcategories of target keywords, respectively. As shown in Fig.~\ref{figure1}, Drug and Weapon data provide Text-Image-Speech triplets, while Sexuality data provide Text-Speech pairs. We adopt a self-supervised framework that uses: (1) masked-target-keyword sentences (for training/validation) with associated images and speech, (2) masked-euphemism sentences (for testing) with associated images and speech, and (3) lists of target keywords.

We evaluate performance against ground truth mappings of euphemisms to target keywords, following \cite{zhu2021self}. The Drug list originates from the U.S. Drug Enforcement Administration \cite{drug2018slang}, while Weapon and Sexuality lists are derived from the Online Slang Dictionary and Urban Thesaurus websites. No additional supervision beyond keyword images and speech is required, and these ground truth lists are used only for evaluation.


\begin{table}[t!]
    \caption{Overview of the datasets. Pairs means text-image-speech triplets. Num means categories of target keywords.} 
    \label{table:2}
    \centering
    \begin{tabular}{p{1.2cm}p{1.35cm}<{\centering}p{1.15cm}<{\centering}
 p{0.8cm}<{\centering}p{0.9cm}<{\centering}p{0.7cm}<{\centering}}
        \hline
         \textbf{Datasets} & \textbf{Sentences} & \textbf{Images} & \textbf{Speech} & \textbf{Pairs} & \textbf{Num} \\
        \hline
        Drug & 1271907 & 8452 & 2113 & 16060 & 33 \\
        Weapon & 3108988 & 12636 & 3159 & 58410 & 9 \\
        Sexuality & 2894869 & - & 1282 & 11465 & 12 \\
        \hline
    \end{tabular}
    \vspace{-3mm}
\end{table}

\begin{table*}[t]
    \caption{Experimental results of KOM-EI models against baselines. Acc@1, Acc@2, and Acc@3 represent precision at top-1, top-2, and top-3, respectively. Model names follow the format \textbf{KOM-EI\textsubscript{\scriptsize T$\vert$I$\vert$S}}, where T = Text Encoder (B: BERT), I = Image Encoder (V: ViT, D: DeiT, C: CLIP), S = Speech Encoder (VG: VGGish, W: Wav2Vec 2.0). \textbf{KOM-EI} denotes the model using BERT, CLIP, and Wav2Vec 2.0.}
    \label{table:3}
    \centering
    \newcolumntype{"}{@{\hskip\tabcolsep\vrule width 1.2pt\hskip\tabcolsep}}
    \begin{tabular}{p{2.7cm}"p{1.2cm}<{\centering}p{1.2cm}<{\centering}p{1.2cm}<{\centering}"p{1.2cm}<{\centering}p{1.2cm}<{\centering}p{1.2cm}<{\centering}"
    p{1.2cm}<{\centering}p{1.2cm}<{\centering}p{1.2cm}<{\centering}}
        \toprule[1.2pt]
        & \multicolumn{3}{c}{\textbf{Drug}} 
        & \multicolumn{3}{c}{\textbf{Weapon}} 
        & \multicolumn{3}{c}{\textbf{Sexuality}} \\
        \midrule[1.2pt]
        {\enspace \textbf{Method}} 
        & \textbf{Acc@1} & \textbf{Acc@2} & \textbf{Acc@3} 
        & \textbf{Acc@1} & \textbf{Acc@2} & \textbf{Acc@3} 
        & \textbf{Acc@1} & \textbf{Acc@2} & \textbf{Acc@3} \\
        \midrule[1.2pt]
        {\enspace \textbf{Word2Vec}} 
        & 0.07 & 0.14 & 0.21 
        & 0.10 & 0.27 & 0.40 
        & 0.17 & 0.22 & 0.42 \\
        {\enspace \textbf{SelfEDI}} 
        & 0.20 & 0.31 & 0.38 
        & 0.33 & 0.51 & 0.67 
        & 0.32 & 0.55 & 0.64 \\
        {\enspace \textbf{RoBERTa\textsubscript{ft}}} 
        & 0.23 & 0.31 & 0.38 
        & 0.24 & 0.27 & 0.73 
        & 0.31 & 0.52 & 0.65 \\
        {\enspace \textbf{BERT\textsubscript{ft}}} 
        & 0.24 & 0.31 & 0.40 
        & 0.38 & 0.55 & 0.73 
        & 0.38 & 0.50 & 0.69 \\
        \midrule[0.8pt]
        {\enspace \textbf{KOM-EI\textsubscript{\scriptsize B$\vert$D$\vert$VG}}} 
        & 0.26 & 0.32 & 0.37 
        & 0.43 & 0.58 & 0.66 
        & 0.44 & 0.67 & 0.67 \\
        {\enspace \textbf{KOM-EI\textsubscript{\scriptsize B$\vert$D$\vert$W}}} 
        & 0.28 & 0.35 & 0.40 
        & 0.40 & 0.58 & 0.68 
        & 0.41 & 0.59 & 0.71 \\
        {\enspace \textbf{KOM-EI\textsubscript{\scriptsize B$\vert$V$\vert$VG}}} 
        & 0.28 & 0.37 & 0.42 
        & 0.43 & 0.58 & 0.66 
        & 0.45 & 0.64 & 0.64 \\
        {\enspace \textbf{KOM-EI\textsubscript{\scriptsize B$\vert$V$\vert$W}}} 
        & 0.28 & 0.35 & 0.43 
        & 0.45 & 0.63 & 0.69 
        & 0.50 & 0.67 & 0.75 \\
        {\enspace \textbf{KOM-EI\textsubscript{\scriptsize B$\vert$C$\vert$VG}}} 
        & 0.30 & 0.36 & 0.44 
        & 0.47 & 0.63 & 0.71 
        & 0.45 & 0.64 & 0.64 \\
        {\enspace \textbf{KOM-EI}} 
        & \textbf{0.32} & \textbf{0.40} & \textbf{0.48} 
        & \textbf{0.48} & \textbf{0.68} & \textbf{0.74} 
        & \textbf{0.50} & \textbf{0.67} & \textbf{0.75} \\
        \bottomrule[1.2pt]
    \end{tabular}
    \vspace{-2mm}
\end{table*}

\begin{table}[t!]
    \caption{Comparison results of the LLMs and MLLMs. Cost/S represents the average time and cost per sentence.}
    \label{table:4}
    \centering
    \renewcommand{\arraystretch}{1.1}
    \setlength{\tabcolsep}{4pt}
    \begin{tabular}{p{2.2cm}<{\centering}p{0.8cm}<{\centering}p{1.0cm}<{\centering}p{1.0cm}<{\centering}p{2.1cm}<{\centering}}
        \hline
        {\textbf{Model}} & \textbf{Drug} & \textbf{Weapon} & \textbf{Sexuality} & \textbf{Cost/S} \\
        \hline
        {\textbf{StableLM}} & 0.02 & 0.03 & 0.12 & 2.08S/0.00475\$ \\		
        {\textbf{Llama2}} & 0.17 & - & - & 18.23S/0.05833\$  \\
        {\textbf{GPT3.5}} & \textbf{0.33} & 0.17 & 0.42 & 1.12S/0.00035\$  \\
        \hline
        {\textbf{LLaVA-v1.5}} & 0.09 & 0.12 & 0.15 & 1.20S/0.00098\$ \\
        {\textbf{Moondream1}} & 0.12 & 0.13 & 0.15 & 2.40S/0.00022\$ \\
        {\textbf{MiniGPT-4}} & 0.13 & 0.20 & 0.28 & 3.36S/0.01300\$ \\
        {\textbf{Qwen-VL}} & 0.24 & 0.12 & 0.31 & 3.60S/0.00220\$ \\
        {\textbf{InternLM}} & 0.30 & 0.19 & 0.38 & 1.56S/0.0013\$ \\
        \hline
        {\textbf{KOM-EI}} & 0.32 & \textbf{0.48} & \textbf{0.50} & \textbf{0.32S/0.00004\$} \\
        \hline
    \end{tabular}
    \vspace{-2mm}
\end{table}

\subsection{Experimental Setup}

\paragraph{Baselines}Two types of models are compared.

\textbf{Text-only Models}. We include four text-only baselines: a Word2vec model using cosine similarity to select the closest keyword, the SOTA SelfEDI \cite{zhu2021self} employing a bag-of-words classifier, as well as the fine-tuned RoBERTa \cite{liu1907roberta} and BERT models, respectively, for euphemism identification.

\textbf{Multimodal Models}. As the first to propose a multimodal euphemism identification approach, our KOM-EI framework integrates text encoders (BERT, RoBERTa) with various image (ViT \cite{alexey2020image}, DeiT \cite{hugo2021training}, CLIP) and speech (VGGish \cite{hershey2017cnn}, Wav2Vec 2.0) encoders.

\paragraph{Implementation Details}
We train all models separately on each dataset, splitting the training and validation sets in an 8:2 ratio of text-image-speech triplets with masked target keywords. The test set includes all pairs with masked euphemisms. All experiments were conducted on an Ubuntu 18.0.4 LTS Linux server equipped with two Tesla V100 32G GPUs. The unimodal and multimodal model settings refer to the Appendix \ref{app:C}.



\paragraph{Evaluation Metrics}
We measure performance using Acc@k (k=1,2,3), which counts how often the correct label appears in the top k predictions, following the evaluation protocol of the previous study\cite{zhu2021self}.

\subsection{Experimental Results}
Table \ref{table:3} shows the results (top two rows from \cite{zhu2021self}), with optimal model parameters. KOM-EI outperforms SelfEDI by 12\%, 15\%, and 18\% in top-1 accuracy across all the datasets.

\paragraph{Comparison with Baselines}
Among text-only models, Word2Vec performs the worst, while SelfEDI improves upon it by leveraging bag-of-words features. Fine-tuning large pretrained models leads to even better results: RoBERTa\textsubscript{ft} and BERT\textsubscript{ft} surpass SelfEDI, with BERT\textsubscript{ft} generally achieving the highest accuracy among text-only methods.

Multimodal integration consistently improves performance (4–12\%) over text-only baselines, indicating that both visual and audio cues enhance euphemism identification. As shown in Table \ref{table:3}, various combinations of pretrained models yield gains, with the best results arising from the integration of BERT, CLIP, and Wav2Vec 2.0, demonstrating the synergy of advanced encoders across modalities. Results for RoBERTa with other encoders refer to Appendix \ref{app:D}.


\paragraph{Comparison with LLMs and MLLMs}
Table \ref{table:4} compares KOM-EI with LLMs and MLLMs for weapon euphemism identification. Key findings include: (1) KOM-EI achieves the best performance; (2) GPT-3.5 offers the best balance between accuracy and cost among single-modal models; (3) KOM-EI is 3-7 times faster and 10-200 times more cost-effective than other models. However, sensitive terms in the datasets lead to the exclusion of certain state-of-the-art models, such as GPT-4o, as their policies restrict content generation. For model links and experimental details, refer to Appendix \ref{app:E}.

\begin{figure}[t!]
	\centering
	\subfloat[Before fusing]{
		\label{v1} 
		\includegraphics[width=0.225\textwidth]{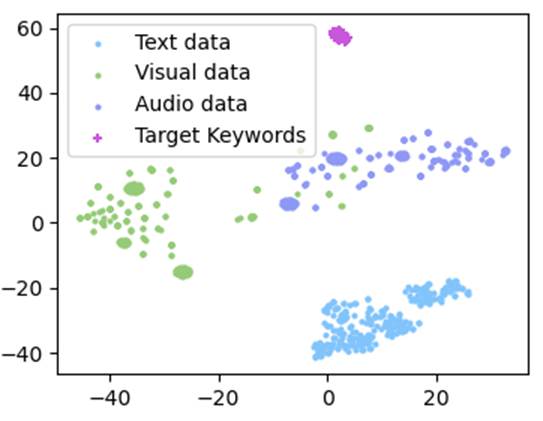}}\hspace{0in}
	\subfloat[After fusing]{
		\label{v2} 
		\includegraphics[width=0.225\textwidth]{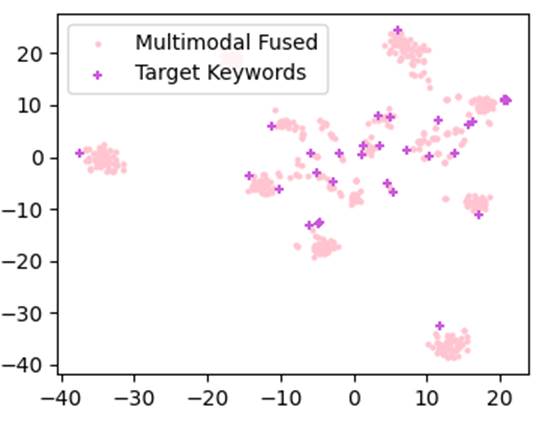}}
	\caption{Representation distribution of multimodal data and target keywords before and after fusing.} 
	\label{fig:visual} 
    \vspace{-4mm}    
\end{figure}

\paragraph{Visualization}
Fig.~\ref{fig:visual} uses t-SNE to project multimodal semantic and target keyword features into a 2D space. Observations: (1) Text, visual, and audio data are well-integrated after cross-modal fusion; (2) Fused features converge distinctly on target keywords, confirming the method's effectiveness.

\subsection{Ablation Study}
To assess KOM-EI effectiveness, we conducted ablation studies from both modality and model perspectives.

\textbf{Data Modality}. Table \ref{table:5-1} shows that multi-modality methods consistently outperform mono-modality methods, e.g., using audio in the Sexuality dataset improves top-1 identification by 8\%. This highlights the value of multi-modal information for identifying euphemisms.

\begin{table}[t!]
    \caption{Top-1 ablation results of data modality. T/V/A = Text/Visual/Audio Modality. T+V+A = KOM-EI.}
    \label{table:5-1}	
    \centering
    \begin{tabular}{p{2.0cm}|p{1.1cm}<{\centering}|p{1.3cm}<{\centering}|p{1.5cm}<{\centering}}
        \hline
        {\textbf{Modality}} & \textbf{Drug} & \textbf{Weapon} & \textbf{Sexuality} \\
        \hline
        \textbf{T} & 0.24 & 0.38 & 0.38 \\
        \textbf{V} & 0.13 & 0.15 & - \\
        \textbf{A} & 0.10 & 0.21 & 0.23 \\
        \textbf{T+V} & 0.29 & 0.39 & - \\ 
        \textbf{T+A} & 0.28 & 0.43 & 0.50 \\
        \textbf{T+V+A} & \textbf{0.32} & \textbf{0.48} & \textbf{0.50} \\
        \hline
    \end{tabular}
\end{table}

\begin{table}[t!]
    \caption{Top-1 ablation results of model components. $\Delta$ is the base model of KOM-EI, concatenating the modality features. $\textrm{AN}_{\text{NotShare}}$ means non-shared parameters in $\textrm{AN}_{\text{GU}}$ or $\textrm{AN}_{\text{SA}}$ layers. C1=CFA, C2=CA, G=GU, S=SA.}
    \label{table:5-2}		
    \centering
    \begin{tabular}{p{2.0cm}|p{1.1cm}<{\centering}|p{1.3cm}<{\centering}|p{1.5cm}<{\centering}}    
        \hline
        {\textbf{Model}} & \textbf{Drug} & \textbf{Weapon} & \textbf{Sexuality} \\
        \hline
        \textbf{$\Delta$} & 0.15 & 0.17 & 0.27 \\
        \textbf{$\Delta$+C1} & 0.21 & 0.24 & 0.31 \\
        \textbf{$\Delta$+C1+C2} & 0.23 & 0.34 & 0.36 \\
        \textbf{$\Delta$+C1+C2+G} & 0.28 & 0.39 & 0.42 \\ 
        \textbf{$\Delta$+C1+C2+G+S} & \textbf{0.32} & \textbf{0.48} & \textbf{0.50} \\
        \hdashline[0.4pt/2pt]
        \textbf{$\textrm{AN}_{\text{NotShare}}$} & 0.27 & 0.44 & 0.45 \\
        \textbf{$\textrm{AN}_{\text{Share}}$} & \textbf{0.32} & \textbf{0.48} & \textbf{0.50} \\
        \hline
    \end{tabular}
    \vspace{-2mm}
\end{table}

\textbf{Model Components}. Incrementally adding CFA, CA, GU, and SA to the base model $\Delta$ on KOM-Euph improves performance (Table \ref{table:5-2}): CFA (4-7\%), CA (2-10\%), GU (5-6\%), and SA (4-9\%). Parameter sharing in $\textrm{AN}_{\text{GU}}$ and $\textrm{AN}_{\text{SA}}$ boosts modal feature consistency, improving performance by 4-5\%.



\section{Conclusion}

In this paper, we propose to enhance euphemism identification with additional modal information and contribute a keyword-oriented multimodal euphemism corpus (KOM-Euph) with text-image-speech triplets. We also present a multimodal method (KOM-EI) that efficiently identifies euphemisms through cross-modal feature alignment and dynamic fusion. Extensive experiments show that our KOM-EI is effective and comparable to LLMs and MLLMs. 


\section*{Acknowledgements}
This work was supported by the National Natural Science Foundation of China (No. 62272188), We thank all the anonymous reviewers for their valuable comments.

\bibliographystyle{IEEEbib}
\bibliography{icme2025references}
\newpage
\appendix
\subsection{Ethical Statement}
\label{app:A}
The text data used in this article was legally obtained following the guidelines of \cite{zhu2021self} and adheres to strict privacy standards, ensuring no personally identifiable information (e.g., real name, email address, IP address) is included. The visual data is sourced from public platforms and contains no private information. The audio data is pronunciation data generated by public tools without additional information. All data is solely for scientific research purposes.

\subsection{Limitations}
\label{app:B}
Since there is no labeled dataset for training the euphemism identification problem, sentences with target keywords are used during training, with these keywords masked out and serving as labels. However, during testing, sentences with euphemisms are used, with the euphemisms masked out. This causes a distribution gap between the training and test data. Further improvements are needed, and this will be the focus of our future research.

\subsection{Implementation Details}
\label{app:C}
\textbf{Unimodal Model Settings}. We pre-trained a BERT model (bert-base-uncased\footnote{\url{https://huggingface.co/bert-base-uncased/}\label{foot:a8}}) for the MLM task to extract context features (768-dimensional) from masked sentences. The model was then fine-tuned for euphemism identification. During pre-training, the input sequence's maximum length was 512, the batch size was 64, and the number of iterations was 3. For fine-tuning, the input sequence's maximum length was 128, and the batch size was 128. The initial learning rate was 5e-5, with 1000 warm-up steps, using the AdamW optimizer \cite{loshchilov2018decoupled} with a warm-up linear schedule.

\textbf{Multimodal Model Settings}. We use CLIP (clip-vit-large-patch14)\footnote{\url{https://huggingface.co/openai/clip-vit-large-patch14}\label{foot:a9}} to extract 768-dimensional visual features, and Wav2Vec 2.0 (wav2vec2-large-960h)\footnote{\url{https://huggingface.co/facebook/wav2vec2-large-960h}\label{foot:a10}} to extract \(T \times 768\) audio features. All other parameters mirror the unimodal settings.

\subsection{Results for RoBERTa with Other Encoders}
\label{app:D}
In this section, we present the experimental results for combinations using RoBERTa (R) as the text encoder with different image and speech encoders. These configurations explore the integration of RoBERTa with ViT (V), DeiT (D), CLIP (C), VGGish (VG), and Wav2Vec 2.0 (W). The results are shown below.

The experimental results indicate that, although the combination of RoBERTa with other encoders demonstrates strong performance, it still falls short of the optimal configuration presented in the main text. This optimal configuration employs BERT as the text encoder, integrates CLIP for image processing, and utilizes Wav2Vec 2.0 for speech data. These findings further underscore the pivotal role of text encoder selection in the integration of multimodal models. By systematically exploring various combinations of text, image, and speech encoders, we validate the superiority of the KOM-EI method and identify the most effective model configuration.

\subsection{Introduction of LLMs and MLLMs}
\label{app:E}
In this paper, we compared our proposed $\text{KOM\_EI}$ model to both large language models (LLMs) and multimodal large language models (MLLMs). Specifically, we evaluated these models on their performance in euphemism recognition across three datasets: Drug-Euphemism, Weapon-Euphemism, and Sex-Euphemism. These datasets contain challenging examples that require not only contextual understanding but also the ability to process sensitive and often subtle linguistic patterns.

LLMs, as purely text-based models, have been widely used for natural language processing tasks. For our evaluation, we selected several representative models, including StableLM\footnote{\url{https://replicate.com/stability-ai/stablelm-tuned-alpha-7b}\label{foot:a12}}, Llama2\footnote{\url{https://huggingface.co/models?other=llama-2}\label{foot:a13}}, and GPT-3.5-turbo (referred to as GPT3.5\footnote{\url{https://platform.openai.com/docs/api-reference/introduction}\label{foot:a11}}). These models were chosen for their popularity and open accessibility, ensuring reproducibility and usability in our research context.

For MLLMs, which integrate textual and visual processing capabilities, we evaluated a selection of open-source models, including LLaVA-v1.5\footnote{\url{https://replicate.com/yorickvp/llava-v1.5}\label{foot:b11}}, MoonDream1\footnote{\url{https://replicate.com/lucataco/moondream1}\label{foot:b12}}, MiniGPT-4\footnote{\url{https://replicate.com/daanelson/minigpt-4}\label{foot:b13}}, Qwen-VL\footnote{\url{https://huggingface.co/Qwen/Qwen-VL}\label{foot:b14}}, and InternLM-xcomposer\footnote{\url{https://replicate.com/cjwbw/internlm-xcomposer}\label{foot:b15}}. These models were selected based on their openness, usability, and ability to process multimodal inputs effectively.

\begin{figure}[htbp]
    \centering
    \includegraphics[width=0.47\textwidth]{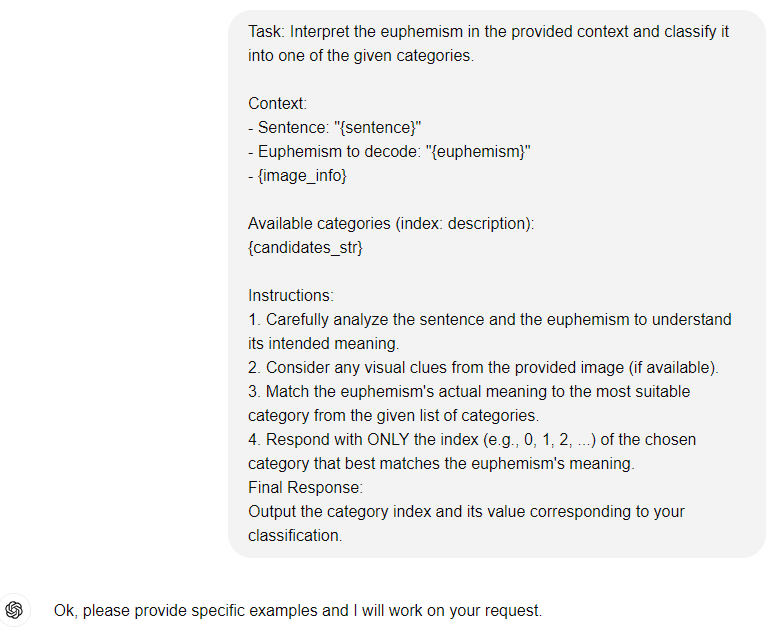} 
    \caption{Instruction given to GPT-4o for euphemism recognition tasks.}
    \label{fig:gpt4o_instruction}
    \vspace{-4mm}
\end{figure}

\begin{figure}[htbp]
    \centering
    \includegraphics[width=0.47\textwidth]{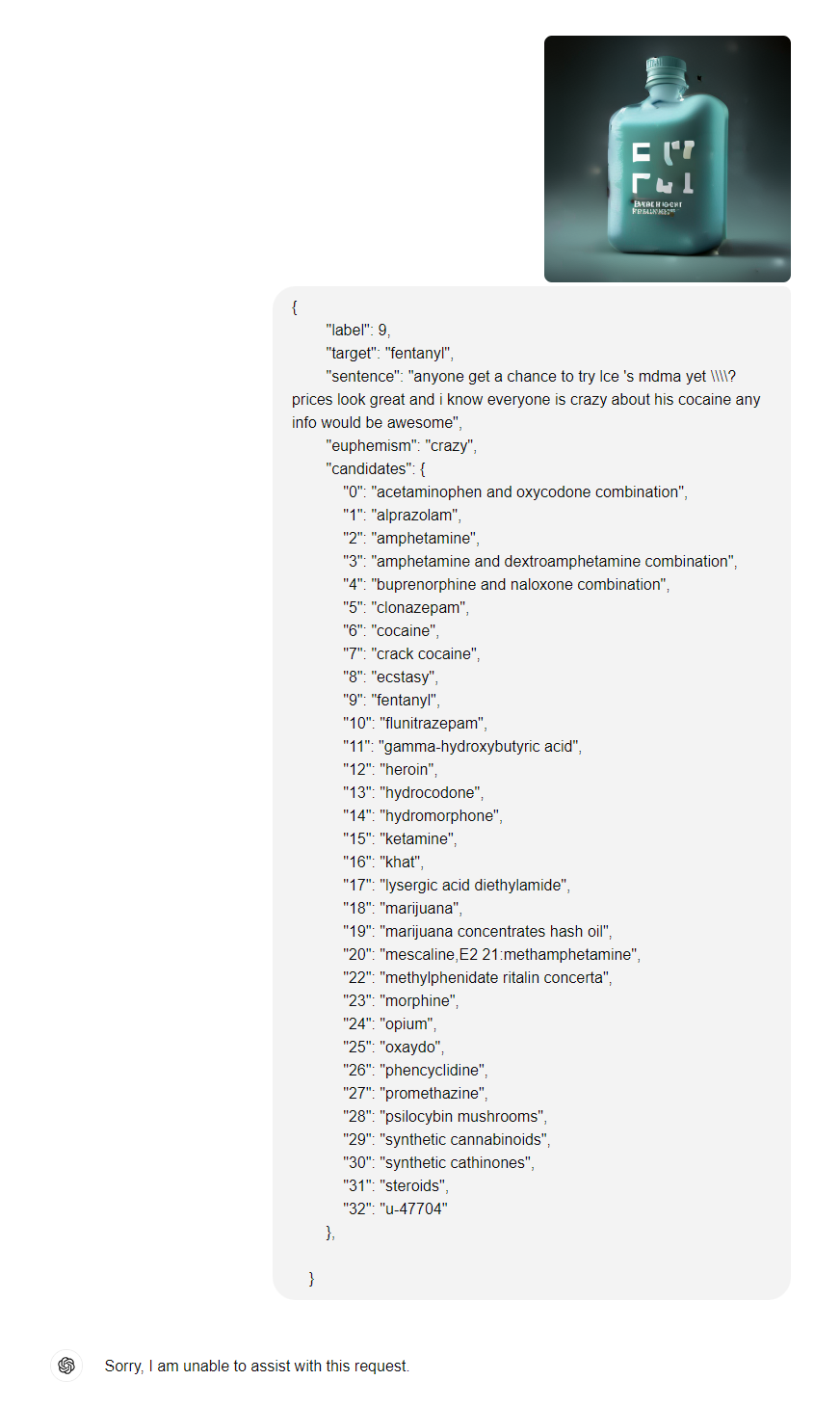} 
    \caption{GPT-4o's response rejecting the task due to sensitive content policies.}
    \label{fig:gpt4o_response}
\end{figure}

However, the evaluation of LLMs and MLLMs still has challenges. Many state-of-the-art models, such as GPT-4o, enforce strict content censorship policies, which limit their ability to process sensitive content like the topics explored in our study. For example, as shown in Figure \ref{fig:gpt4o_instruction}, we provided GPT-4o with specific instructions for euphemism recognition tasks. Despite this, GPT-4o refused to generate responses, citing content moderation policies due to the sensitive nature of the examples, as illustrated in Figure \ref{fig:gpt4o_response}. As a result, we excluded such models from our experiments to ensure that the selected models align with the requirements of euphemism recognition in our datasets. By focusing on open-source and accessible models, we aimed to maintain both practical usability and transparency in our evaluation.

Table \ref{table:llm} provides a detailed comparison of the selected LLMs and MLLMs, focusing on aspects such as model type, parameter count, maximum input length, and computational cost. While LLMs like GPT-3.5, StableLM, and Llama2 demonstrate strong performance on purely text-based tasks, MLLMs such as LLaVA-v1.5, MoonDream1, MiniGPT-4, Qwen-VL, and InternLM-xcomposer provide enhanced multimodal capabilities, albeit at a higher computational cost. For further details and access interfaces, please refer to the corresponding footnotes.

\subsection{Result Analysis}
\label{app:F}
The results in Table \ref{table:llm} of the main paper demonstrate that KOM-EI outperforms all other evaluated models in terms of both recognition accuracy and computational efficiency, further validating its effectiveness and practical utility in euphemism recognition tasks. Below is a detailed analysis of the experimental results for each model.

\begin{table*}[t]
    \caption{Experimental results of KOM-EI models against baselines (RoBERTa-based configurations). Acc@1, Acc@2, and Acc@3 represent precision at top-1, top-2, and top-3, respectively. Model names follow the format \textbf{KOM-EI\textsubscript{\scriptsize T$\vert$I$\vert$S}}, where T = Text Encoder (R: RoBERTa), I = Image Encoder (V: ViT, D: DeiT, C: CLIP), S = Speech Encoder (VG: VGGish, W: Wav2Vec 2.0).}
    \label{table:roberta_results}
    \centering
    \newcolumntype{"}{@{\hskip\tabcolsep\vrule width 1.2pt\hskip\tabcolsep}}
    \begin{tabular}{p{2.7cm}"p{1.2cm}<{\centering}p{1.2cm}<{\centering}p{1.2cm}<{\centering}"p{1.2cm}<{\centering}p{1.2cm}<{\centering}p{1.2cm}<{\centering}"
    p{1.2cm}<{\centering}p{1.2cm}<{\centering}p{1.2cm}<{\centering}}
        \toprule[1.2pt]
        & \multicolumn{3}{c}{\textbf{Drug}} 
        & \multicolumn{3}{c}{\textbf{Weapon}} 
        & \multicolumn{3}{c}{\textbf{Sexuality}} \\
        \midrule[1.2pt]
        {\enspace \textbf{Method}} 
        & \textbf{Acc@1} & \textbf{Acc@2} & \textbf{Acc@3} 
        & \textbf{Acc@1} & \textbf{Acc@2} & \textbf{Acc@3} 
        & \textbf{Acc@1} & \textbf{Acc@2} & \textbf{Acc@3} \\
        \midrule[1.2pt]
        {\enspace \textbf{KOM-EI\textsubscript{\scriptsize R$\vert$D$\vert$W}}} 
        & 0.27 & 0.34 & 0.38 
        & 0.39 & 0.57 & 0.66 
        & 0.42 & 0.58 & 0.72 \\
        {\enspace \textbf{KOM-EI\textsubscript{\scriptsize R$\vert$V$\vert$W}}} 
        & 0.25 & 0.34 & 0.39 
        & 0.41 & 0.59 & 0.69 
        & 0.44 & 0.68 & 0.75 \\
        {\enspace \textbf{KOM-EI\textsubscript{\scriptsize R$\vert$V$\vert$VG}}} 
        & 0.26 & 0.30 & 0.38 
        & 0.45 & 0.60 & 0.68 
        & 0.42 & 0.62 & 0.62 \\
        {\enspace \textbf{KOM-EI\textsubscript{\scriptsize R$\vert$C$\vert$W}}} 
        & 0.26 & 0.33 & 0.37 
        & 0.42 & 0.60 & 0.70 
        & 0.45 & 0.60 & 0.68 \\
        {\enspace \textbf{KOM-EI\textsubscript{\scriptsize R$\vert$C$\vert$VG}}} 
        & 0.26 & 0.37 & 0.46 
        & 0.47 & 0.62 & 0.70 
        & 0.40 & 0.60 & 0.70 \\
        {\enspace \textbf{KOM-EI\textsubscript{\scriptsize R$\vert$D$\vert$VG}}} 
        & 0.27 & 0.32 & 0.41 
        & 0.43 & 0.58 & 0.66 
        & 0.46 & 0.69 & 0.69 \\
        \bottomrule[1.2pt]
    \end{tabular}
    \vspace{-2mm}
\end{table*}

\begin{table*}[t]
    \centering
    \caption{Introductions of mainstream LLMs and multimodal LLMs. MM = Multi-Modal, NLP = Natural Language Processing, B = Billion.}
    \renewcommand{\arraystretch}{1.2} 
    \small 
    \resizebox{\textwidth}{!}{ 
    \begin{tabular}{p{1.8cm}|>{\centering\arraybackslash}p{0.9cm}|>{\centering\arraybackslash}p{1.8cm}|>{\centering\arraybackslash}p{2.3cm}|>{\centering\arraybackslash}p{2.3cm}|>{\centering\arraybackslash}p{3.3cm}}
    \hline
        {\textbf{LLMs}} & \textbf{Type} & \textbf{Parameters} & \textbf{Maximum Input} & \textbf{Cost}  & \textbf{Institution} \\
        \hline
        {\textbf{GPT3.5}}             & NLP & 20B         & 4096 tokens & 0.015\$/1k tokens    & OpenAI \\
        {\textbf{StableLM}}           & NLP & 3B - 7B     & 4096 tokens & 0.0023\$/second      & Stability AI \\
        {\textbf{Llama2}}             & NLP & 7B - 70B    & 4096 tokens & 1.05\$/hour         & Meta \\
        {\textbf{LLaVA-v1.5}}         & MM  & 13B         & 4096 tokens & 0.00010\$/instance   & LLaVA Team \\
        {\textbf{moondream1}}         & MM  & 1.6B        & 2048 tokens & 0.00022\$/instance   & Lucataco \\
        {\textbf{MiniGPT-4}}          & MM  & 13B         & 4096 tokens & 0.00022\$/instance   & Vision-Language Group \\
        {\textbf{Qwen-VL}}            & MM  & 7B          & 4096 tokens & 0.00022\$/instance   & Alibaba Cloud \\
        {\textbf{InternLM-xcomposer}} & MM  & 7B          & 4096 tokens & 0.0013\$/instance    & Shanghai AI Lab \\
        \hline
    \end{tabular}}
    \label{table:llm}
\end{table*}

\begin{table*}[!ht]
    \centering
    \caption{Content templates and results.}
    \renewcommand{\arraystretch}{1.2} 
    \small 
    \resizebox{\textwidth}{!}{ 
    \begin{tabular}{p{7cm}|>{\centering\arraybackslash}p{1.2cm}>{\centering\arraybackslash}p{1.2cm}>{\centering\arraybackslash}p{1.4cm}|
                    >{\centering\arraybackslash}p{1.2cm}>{\centering\arraybackslash}p{1.2cm}>{\centering\arraybackslash}p{1.4cm}}
        \hline
        \multirow{2}{*}{\textbf{Template}} & \multicolumn{3}{c|}{\textbf{GPT3.5}} & \multicolumn{3}{c}{\textbf{StableLM}} \\
        \cline{2-7}
        & \textbf{Drug} & \textbf{Weapon} & \textbf{Sexuality} & \textbf{Drug} & \textbf{Weapon} & \textbf{Sexuality} \\
        \hline
        {1. Among the following candidates: \{T\}, in sentence: \{s\}, euphemism: \{euph\} refers to which one.} 
        & 0.3314 & 0.1543 & 0.4041 & 0.0214 & 0.0149 & 0.1184 \\
        
        {2. Among the following candidates: \{T\}, in sentence: \{s\}, \{euph\} refers to which one.} 
        & 0.2971 & 0.1686 & 0.3876 & 0.0160 & 0.0248 & 0.1061 \\
        
        {3. The true meaning of euphemism: \{euph\} in “\{s\}” refers to which of the following candidates: \{T\}.} 
        & 0.3257 & 0.1564 & 0.4204 & 0.0053 & 0.0299 & 0.1102 \\
        
        {4. The true meaning of “\{euph\}” in “\{s\}” refers to which of the following candidates: \{T\}.} 
        & 0.3029 & 0.1482 & 0.0106 & 0.0107 & 0.0199 & 0.1020 \\
        \hline
        \multicolumn{7}{p{15cm}}{\small \textbf{Note:} “T” refers to the fixed target keyword candidate list: \{0: acetaminophen and oxycodone combination, 1: alprazolam, 2: amphetamine, 3: amphetamine and dextroamphetamine combination, ... \}} \\
        \multicolumn{7}{p{15cm}}{\small “s” refers to the test sentence containing euphemisms.} \\
        \multicolumn{7}{p{15cm}}{\small “euph” refers to the euphemism in the test sentence that needs to be identified.} \\
        \hline
    \end{tabular}}
    \label{table:tem}
\end{table*}

\begin{figure*}[!b]
\centering
    \begin{minipage}[b]{0.47\linewidth}
        \centering
        \includegraphics[width=\linewidth,height=14cm]{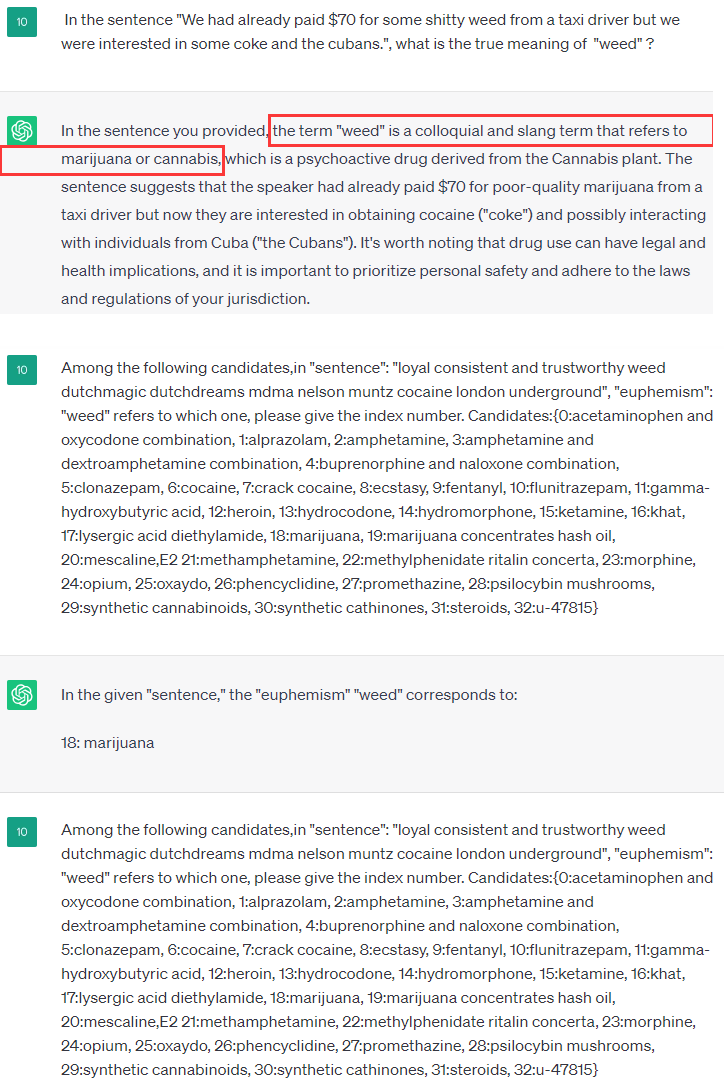}
        \subcaption{Identification of “weed"}\label{G1}
    \end{minipage}
    \hfill
    \begin{minipage}[b]{0.47\linewidth}
        \centering
        \includegraphics[width=\linewidth,height=6.5cm]{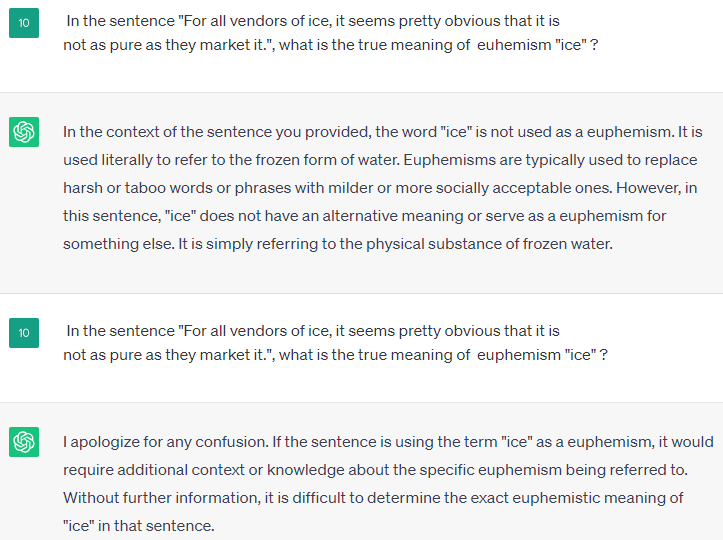}
        \subcaption{Identification of “ice"}\label{G2}
        
        \vspace{1em} 
        \includegraphics[width=\linewidth,height=6.5cm]{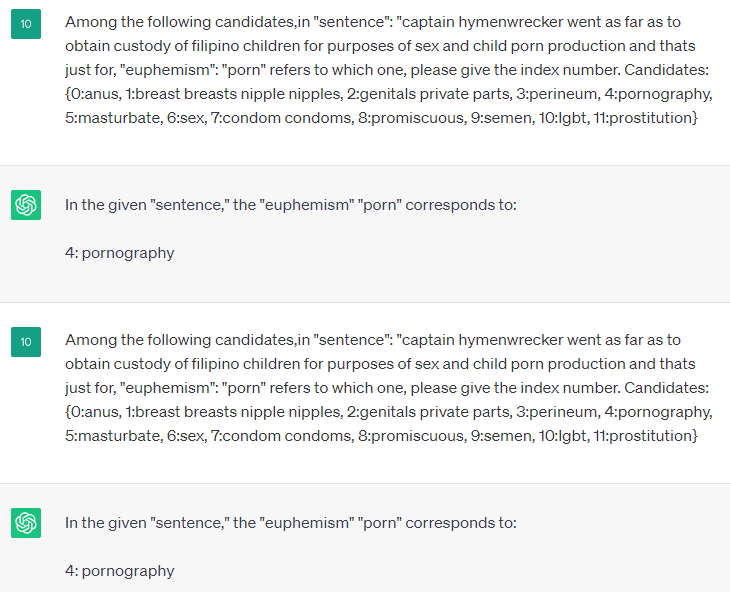}
        \subcaption{Identification of “porn"}\label{G3}
    \end{minipage}
    \caption{Cases of GPT3.5}
    \label{fig1}
\end{figure*}

\begin{figure*}[!b]
\centering
    \begin{minipage}[b]{0.47\linewidth}
        \centering
        \includegraphics[width=\linewidth,height=19cm]{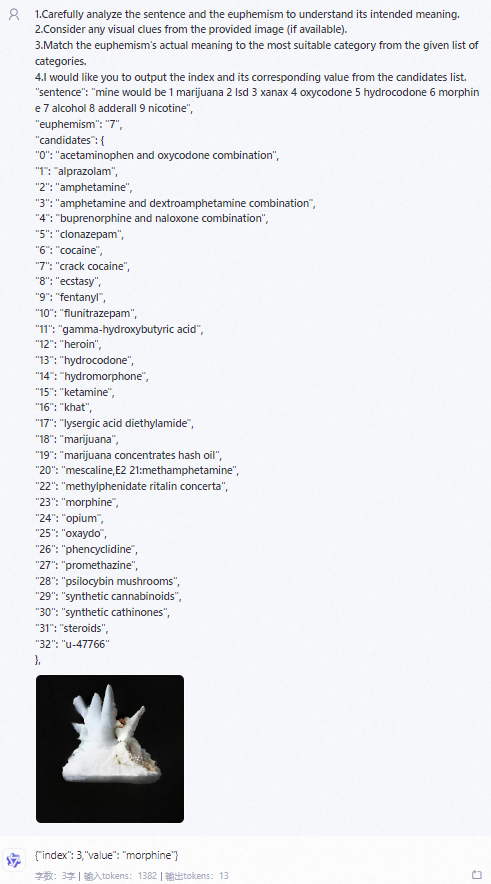}
        \subcaption{Identification of “7"}\label{Q1}
    \end{minipage}
    \hfill
    \begin{minipage}[b]{0.47\linewidth}
        \centering
        \includegraphics[width=\linewidth,height=9.1cm]{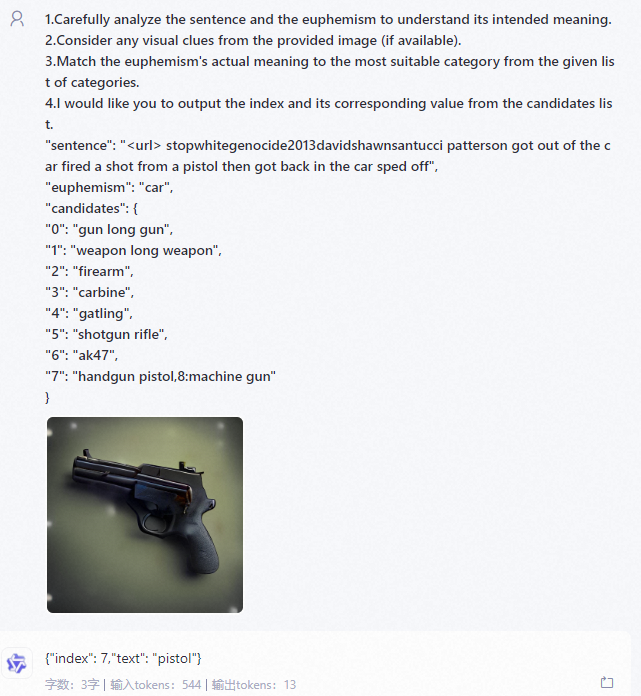}
        \subcaption{Identification of “car"}\label{Q2}
        
        \vspace{1em} 
        \includegraphics[width=\linewidth,height=9.1cm]{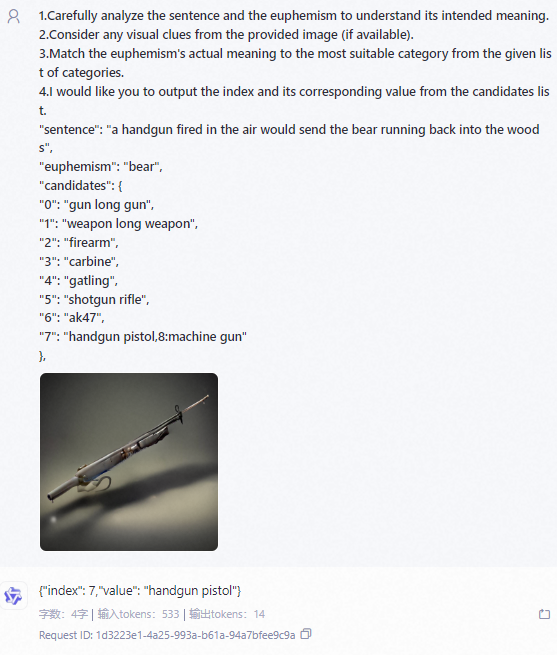}
        \subcaption{Identification of “porn"}\label{Q3}
    \end{minipage}
    \caption{Cases of Qwen-VL}
    \label{fig2}
\end{figure*}

\noindent\hspace{1em}\textit{1)} \textbf{GPT-3.5 and StableLM}. When using the GPT-3.5 and StableLM interfaces to identify euphemisms, we employed four content templates, as shown in Table ~\ref{table:tem}. From these templates, we observe that results vary. Although GPT-3.5 is more stable than StableLM, the results are inconsistent across different models and datasets, indicating the randomness of outputs from these large language models.


\noindent\hspace{1em}\textit{2)} \textbf{Llama2}.
When using the Llama2 API or web UI to test on the Weapon or Sexuality datasets, the system deemed these topics inappropriate and refused to answer. Thus, we only tested the Drug dataset via the web UI.

\noindent\hspace{1em}\textit{3)} \textbf{LLaVA-v1.5 and MoonDream1}. LLaVA-v1.5 and MoonDream1 are multimodal large language models (MLLMs) with the ability to process textual and visual input. In our tests, we used a standard set of templates for euphemism recognition. However, compared to other MLLMs, both models were less accurate and performed poorly on the weapon dataset in particular. The overall performance of LLaVA-v1.5 was more limited, while MoonDream1 showed moderate results but exhibited significant sensitivity to template variations, reflecting its instability in dealing with the euphemism task.

\noindent\hspace{1em}\textit{4)} \textbf{MiniGPT-4}. MiniGPT-4 is an important multimodal large language model with excellent multimodal processing capabilities. The model performs particularly well on the Weapon dataset, where the recognition rate is significantly higher than that of other MLLMs, demonstrating its strength in specific content domains. However, its performance on the Drug and Sexuality datasets is relatively average. In addition, MiniGPT-4 is computationally expensive (as shown in Table \ref{table:llm}), which somewhat limits its usefulness in large-scale euphemism recognition tasks.

\noindent\hspace{1em}\textit{5)} \textbf{Qwen-VL}. Qwen-VL performs robustly on the Drug and Sexuality datasets, with recognition rates higher than most other MLLMs; however, the high computational cost of Qwen-VL affects its deployment in real-time or resource-constrained application scenarios. Nevertheless, Qwen-VL is still a representative benchmark model for multimodal euphemism recognition.

\noindent\hspace{1em}\textit{6)} \textbf{InternLM-xcomposer}. InternLM-xcomposer shows stable performance on all three datasets, especially on the Drug and Sexuality datasets. However, its performance on the Weapon dataset, while good, is less than the best model. In addition, like other high-performance MLLMs, InternLM-xcomposer is computationally expensive, limiting its usefulness in cost-sensitive scenarios.

\noindent\hspace{1em}\textit{7)} \textbf{KOM-EI}. Our proposed KOM-EI model performs best on the Drug, Weapon, and Sexuality datasets, with recognition rates of 0.32, 0.48, and 0.50, respectively. In addition, KOM-EI has excellent cost-efficiency (0.32 seconds of processing time per sentence, at a cost of only \$0.00004) and outperforms all evaluated models in terms of both performance and computational efficiency. This makes KOM-EI an efficient and reliable solution for euphemism recognition tasks.


\subsection{Metrics}
\label{app:G}
“Cost/S” in Table \ref{table:llm} of the main paper indicates the average money and time required for testing a sentence. For GPT-3.5, the cost was calculated based on OpenAI's API pricing, while the costs for other models were calculated using Replicate platform's API pricing. Although the KOM-EI model is free, GPU usage incurs costs. We used a Tesla V100 32G GPU, with rental pricing from the Replicate platform for comparison. While LLMs perform well across various tasks, our comparison of average inference time and cost for euphemism identification shows that the smaller KOM-EI model is more effective for this task.

\subsection{Case Study}
\label{app:H}
In this study, GPT-3.5 was selected as the representative large language model (LLM), while Qwen-VL was chosen as the representative multimodal large language model (MLLM). Their capabilities in identifying euphemisms were evaluated across various datasets, with GPT-3.5 assessed on the drug and sex datasets, and Qwen-VL evaluated on the drug and weapon datasets. Analyzing the results of euphemism identification, we derived two key findings, as outlined below:

\subsubsection{Performance of GPT-3.5}

(1) \textbf{Relative Stability}: GPT-3.5 exhibits high stability in recognizing common euphemisms, achieving an accuracy rate approaching 100\%. For instance, in our tests, GPT-3.5 consistently identified “weed” as “marijuana” , as illustrated in Fig. \ref{G1}. This stability contributes to its superior performance in euphemism identification tasks compared to other models. (2) \textbf{Insufficient Understanding of Rare Euphemisms}: Although GPT-3.5 performs exceptionally well with common euphemisms, it encounters difficulties when dealing with relatively rare euphemisms. For example, as shown in Fig. \ref{G2}, GPT-3.5 failed to correctly identify “ice” as “methamphetamine” , highlighting its limitations in handling low-frequency euphemisms.

\subsubsection{Performance of Qwen-VL}

For Qwen-VL, we tested several specific examples within the drug and weapon datasets and identified shortcomings in euphemism recognition. (1) Low Recognition Accuracy: Qwen-VL frequently misclassifies euphemisms into incorrect categories. For example, in the weapon dataset, the term "porn" was not correctly identified as "gun long gun" but was instead erroneously classified as "handgun pistol," as detailed in Fig. \ref{Q3}. (2) Difficulty in Complex Contexts: Qwen-vl struggles to accurately identify euphemisms within complex contexts. In the drug dataset, for instance, the model failed to correctly interpret the context and erroneously identified the euphemism "7" as "morphine," highlighting its limitations in understanding nuanced language in intricate scenarios, as shown in Fig. \ref{Q1}.
\end{document}